\documentclass{article}

\usepackage{microtype}
\usepackage{graphicx}
\usepackage{booktabs} 

\usepackage{hyperref}

\usepackage[accepted]{icml2025}

\usepackage{amsmath}
\usepackage{amssymb}
\usepackage{mathtools}
\usepackage{amsthm}
\usepackage{subcaption}

\usepackage[capitalize,noabbrev]{cleveref}

\theoremstyle{plain}

\theoremstyle{definition}

\theoremstyle{remark}

\usepackage[textsize=tiny]{todonotes}

\icmltitlerunning{DoMINO: A Decomposable Multi-scale Iterative Neural Operator}

\begin{document}

\twocolumn[
\icmltitle{DoMINO: A Decomposable Multi-scale Iterative Neural Operator  \\
           for Modeling Large Scale Engineering Simulations}



\icmlsetsymbol{equal}{*}

\begin{icmlauthorlist}
\icmlauthor{Rishikesh Ranade}{comp}
\icmlauthor{Mohammad Amin Nabian}{comp}
\icmlauthor{Kaustubh Tangsali}{comp}
\icmlauthor{Alexey Kamenev}{comp}
\icmlauthor{Oliver Hennigh}{comp}
\icmlauthor{Ram Cherukuri}{comp}
\icmlauthor{Sanjay Choudhry}{comp}
\end{icmlauthorlist}
\icmlaffiliation{comp}{NVIDIA, Santa Clara, United States}

\icmlcorrespondingauthor{Rishikesh Ranade}{rranade@nvidia.com}

\icmlkeywords{Local Learning, Engineering Simulation}

\vskip 0.3in
]



\printAffiliationsAndNotice{}  

\begin{abstract}
Numerical simulations play a critical role in design and development of engineering products and processes. Traditional computational methods, such as CFD, can provide accurate predictions but are computationally expensive, particularly for complex geometries. Several machine learning (ML) models have been proposed in the literature to significantly reduce computation time while maintaining acceptable accuracy. However, ML models often face limitations in terms of accuracy and scalability and depend on significant mesh downsampling, which can negatively affect prediction accuracy and generalization. In this work, we propose a novel ML model architecture, DoMINO (Decomposable Multi-scale Iterative Neural Operator) developed in NVIDIA Modulus to address the various challenges of machine learning based surrogate modeling of engineering simulations. DoMINO is a point cloud-based ML model that uses local geometric information to predict flow fields on discrete points. The DoMINO model is validated for the automotive aerodynamics use case using the DrivAerML dataset. Through our experiments we demonstrate the scalability, performance, accuracy and generalization of our model to both in-distribution and out-of-distribution testing samples. Moreover, the results are analyzed using a range of engineering specific metrics important for validating numerical simulations. 
\end{abstract}

\section{Introduction}
\label{introduction}

Simulation and modeling of engineering applications involve representing complex physical and chemical processes with partial differential equations (PDEs) followed by solving these equations on a computational domain with given boundary conditions. Engineering simulation tools are predominantly used in the development of innovative and optimized designs. Typically, simulations are based on numerical methods such as finite volume or element methods \cite{leveque2002finite, leveque2007finite}, which are used to discretize the computational domain into smaller elements known as a mesh, approximating the various partial differentials and solving the system of equations derived on the mesh to obtain the discrete PDE solutions. In many cases, the choice of discretization is a trade-off between accuracy and computational cost. The computational time and memory grows proportionally with the number of elements as it directly impacts the process of meshing the computational domain as well as solving the discrete equations on this mesh. As a result, there has been a constant drive to develop newer numerical methods that can provide greater accuracy as well as greater computational efficiency, i.e., faster and ideally at a lower computational cost.

A significant amount of work has been carried out in improving the computational efficiency of existing methods, development of new methods \cite{sharma2020current, angel2024immersed, ashton2025high, bernaschi2023multi}, and also by utilizing the latest High Performance Computing (HPC) hardware such as GPUs, arm64-based processors etc \cite{stone2021accelerating, appa2021performance, piscaglia2023gpu, xue2024cpu}. Despite these advancements, there has been a continual need to explore Machine Learning (ML) approaches to serve as surrogate models for numerical simulations and provide further acceleration. ML methods developed for numerical simulations have primarily fallen into 2 categories, neural operators and mesh (or graph-)-based approaches. 

Neural operators are a class of deep-learning methods that learn relationships functional spaces. These are mesh-free, infinite dimensional operators that are learned using neural networks \cite{lu2019deeponet, li2020fourier, li2020multipole, patel2021physics}. Recently, significant advancements have been made in applying these methods to large-scale simulations such as computational fluid dynamcics, weather modeling etc \cite{azizzadenesheli2024neural}. Methods are being developed to learn and represent geometries to develop more accurate and generalizable methods. \citet{li2024geometry} adapted to neural operators to modeling directly on meshes. They learn a graph neural operator to project surface meshes of geometries on to structured latent spaces, apply fourier layers and decode the solutions back on the structured meshes. \citet{zhong2025physics} developed a physics-informed geometry aware neural network where the geometry representations was learned from point cloud representation of the surfaces and combined that with physics losses to improve the robustness of the learning algorithm. \citet{he2024geom, park2024point} used geometry parameters and signed distance field (SDF) information in their DeepONet architecture to model a structural problem.  

Mesh-based methods have been another area of focus in modeling numerical simulations. In these methods, ML models are directly trained on simulation meshes and learn to predict solution fields using local mesh information. Early works in this field, such as GraphNet \cite{allen2023graph} and Gated Graph Neural Networks (GGNNs) \cite{li2015gated}, have been applied to a wide range of tasks, including fluid dynamics and structural mechanics, where they model physical interactions through message passing between mesh nodes. In recent years, these methods have been enhanced in works such as MeshGraphNet \cite{pfaff2020learning, sanchez2020learning}, where the geometries are represented with nodal information and physical relationships are learned using mesh faces to train more accurate and generalizable ML models. While these GNN-based models have made significant advancement in modeling numerical simulations, they face challenges in terms of scalability and capturing long-range interactions. Several methods and frameworks have been developed to address the scalability challenge by decomposing the graphs into smaller subdomains and parallelizing the training algorithms on multi-gpu and multi-node architectures \cite{nabian2024x, kakka2024sampling, nastorg2024scalable}. 

The classes of ML methods described above provide different paradigms for modeling numerical simulations. However, a few shortcomings of these methods are outlined below, and subsequently are motivations for this work. Most ML methods struggle to scale to large-scale simulations with element counts ranging in hundreds of millions or billions due to significant memory and compute requirements. The methods that scale to large meshes demonstrate lower accuracy and generalizability due to their inability to efficiently represent the geometries and capture long-range interactions. Typically, ML methods try to learn global geometry representations which are then used to predict solution fields on the surface and in the volume. However, global geometry representations are high-dimensional and dense and are not efficient in accurately predicting solution fields especially in large computational domains and when geometries have finer features. Finally, many of the ML methods are dependent on the spatial structure of the input and do not generalize to other arbitrary distributions. For example, ML algorithms trained on simulation meshes do not generalize and have lower accuracy when evaluated on uniform point clouds or meshes. 

In this work, we propose a novel model architecture, DoMINO, a decomposable, multi-scale, iterative neural operator, to provide scalable, accurate and generalizable surrogates to model large-scale numerical simulations. We demonstrate the DoMINO model using the external aerodynamics use case. External aerodynamics simulations on realistic car geometries involve significantly large surface and volume meshes to capture the near-wall effects and resolve the flow fields in the downstream wake and near other parts of the car such as underbody, side mirrors etc. Accurate and fast predictions of velocity fields, pressure and wall-shear-stresses obtained from aerodynamic simulations are important for providing designers with guidance in developing optimized and innovative vehicle designs. Due to the large-scale nature of this use case, it is ideal for validating the scalability, accuracy and generalizability of ML models. Additionally, as vehicle geometries are complex it requires ML models to efficiently represent the large and small features present in these geometries to enable accurate predictions. In the past, several studies have explored the development of ML surrogates for this application. \citet{chen20213d} use a 3-D UNet architecture \cite{ronneberger2015u} in combination with a custom loss function including the continuity equation loss to model flow fields around vehicles. They use SDFs as inputs to their model to represent and parameterize the car geometries. \citet{jacob2021deep} follows a similar approach where they use a modified 3-D UNet architecture to learn a mapping between a SDF input and the output flow fields. However, they also use the bottleneck layer of the UNet to simultaneously predict the drag coefficient using a fully-connected neural network. \citet{ananthan2024machine} provide a comprehensive study of modeling vehicle aerodynamics. They provide 4 different case studies, ranging from surface predictions using MeshGraphNets \cite{pfaff2020learning} on the Drivaer car dataset, volume predictions with UNet on a motorbike dataset generated OpenFoam tutorial \cite{jasak2007openfoam} to a stable diffusion model to generate different variations starting from a base car geometry. \citet{elrefaie2024drivaernet} proposed a RegDGCNN model to directly learn surface aerodynamic quantities from meshes by integrating encoding capability of PointNet with graph CNNs. These advances demonstrate the potential of machine learning in accelerating design cycles by providing rapid predictions of aerodynamics quantities. However, challenges still exist in terms of generalizing across diverse geometries, scaling to large simulation meshes and the ability to accurately predict both surface and volume aerodynamic quantities.

In this paper, we demonstrate the DoMINO model for the external aerodynamics use case to address these challenges. We train the model on a select set of samples from the DrivAerML dataset \cite{ashton2024drivaerml} and validate the model on the remaining samples. The DrivAerML dataset is chosen for this experiement because it consists of high-fidelity simulation results on large meshes ranging in hundreds of millions. Moreover, the geometric variations in this dataset are significant, resulting in drag force variations from 300 N to 600 N across the various designs. Through the experiments we demonstrate the ability of the DoMINO model to capture both surface and volume flow fields as well as other relevant engineering metrics such as design trends and drag force comparisons. Additionally, we showcase that DoMINO model is mesh independent by validating the surface predictions on a uniformly sampled point cloud instead of the simulation mesh. The results highlight the potential of DoMINO as a practical
and efficient surrogate models for aerodynamic evaluation, supporting real-time applications in automotive design and optimization.

The paper is structured as follows. In Section \ref{domino}, we provide an overview of the DoMINO model architecture and delve deeper into the details of the architecture. Section \ref{experiments} presents additional details related to the experiment, problem setup, dataset etc. Following this, we present the results and analysis in Section \ref{results}.

\section{DoMINO Model Overview}
\label{domino}
In this section, we introduce the DoMINO approach to learn local geometry encodings from point cloud representations to predict PDE solutions on discrete points sampled in computational domain using dynamically constructed computational stencils in local regions around it. We present how the DoMINO model leverages local features within sub-regions of the computational domain to predict solutions on both the surface of geometry as well as in the computational domain volume. Predicting both sets of quantities is extremely important in industrial applications for making crucial choices and decisions regarding product and process design.

\begin{figure}[ht!]
\begin{center}
\centerline{\includegraphics[width=0.8\linewidth]{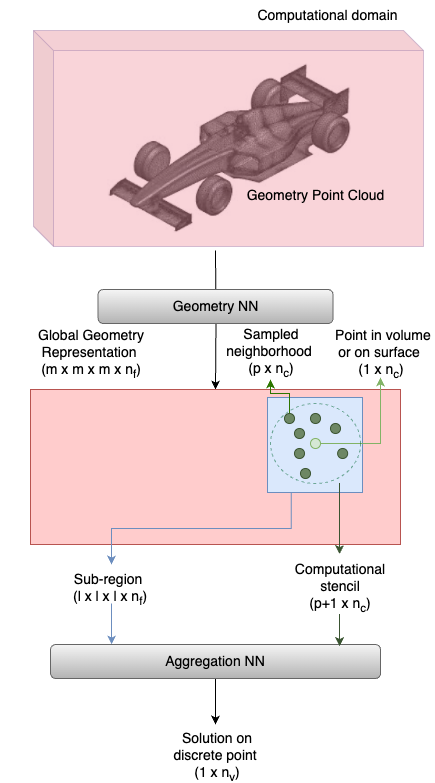}}
\caption{DoMINO architecture}
\label{fig1}
\end{center}
\end{figure}

An overview diagram for the DoMINO model architectures is shown in Fig. \ref{fig1}. Although we present an automotive aerodynamics example for explanation purposes, the approach is applicable to other applications in engineering simulation as well. DoMINO takes the 3-D surface mesh of the geometry as input. A 3-D bounding box is constructed around the geometry to represent a computational domain that captures the most important solutions fields required for design guidance. The arbitrary point cloud representation is transformed into a N-D fixed, structured representation of resolution $m \times m \times m \times f$ defined on the computational domain. A multi-resolution, iterative approach described in Section \ref{global_encoding}, is used to propagate geometry representation into the computational domain and to learn short- and long-range dependencies. The N-D structured representation is a unique global encoding for a given geometry and encapsulates the relevant features of the geometry. However the global encoding is high-dimensional and dense, while the solution at any point in the computational domain is heavily influenced by the information in its locality. As a result, a local encoding needs to be extracted from the global geometry encoding to enable accurate predictions of solutions.

Next, we sample a batch of discrete points in the computational domain where we want to evaluate the solution. When the model is training, these can be sampled from points where the solution is known, for example the nodes of the simulation mesh, while during inference they can be sampled randomly as the simulation mesh may not be available. For each of the sampled points in a batch, a subregion of size $l \times l \times l$ is defined around it and a local geometry encoding is calculated. The local encoding is essentially a subset of the global encoding depending on its position in the computational domain and is calculated using point convolutions. Additional details about the local encoding are provided in  Section \ref{local_encoding}.

Furthermore, for each of the sampled points in a batch $p$ neighboring points are sampled in the computational domain to form a local computational stencil of $p+1$ points. A batch of computational stencils are represented by their local coordinates in the domain and normals whenever available, especially for points sampled on the surface of the geometry. An aggregation network is trained to to predict the solutions on each of the discrete points in the batch using the local geometry encoding and the input features of the computational stencil. More details about the aggregation network are provided in Section \ref{aggregation_net}.

The solution variables on surface typically differ from those calculated in the volume. For example, in the external aerodynamics use case, the solutions variables on surface include pressure and the wall shear stress vector and the volume variables include pressure, velocity and turbulence parameters. As a result, the aggregation neural network needs to be defined separately for predicting volume and surface variables but the global geometry encoding network can be shared between them.
The code for the DoMINO is available in the Modulus repository \cite{contributors2023nvidia} on Github (\url{https://github.com/NVIDIA/modulus/tree/main/examples/cfd/external_aerodynamics/domino}).

\subsection{Global geometry representation}
\label{global_encoding}

\begin{figure}[ht!]
\begin{center}
\centerline{\includegraphics[width=0.8\columnwidth]{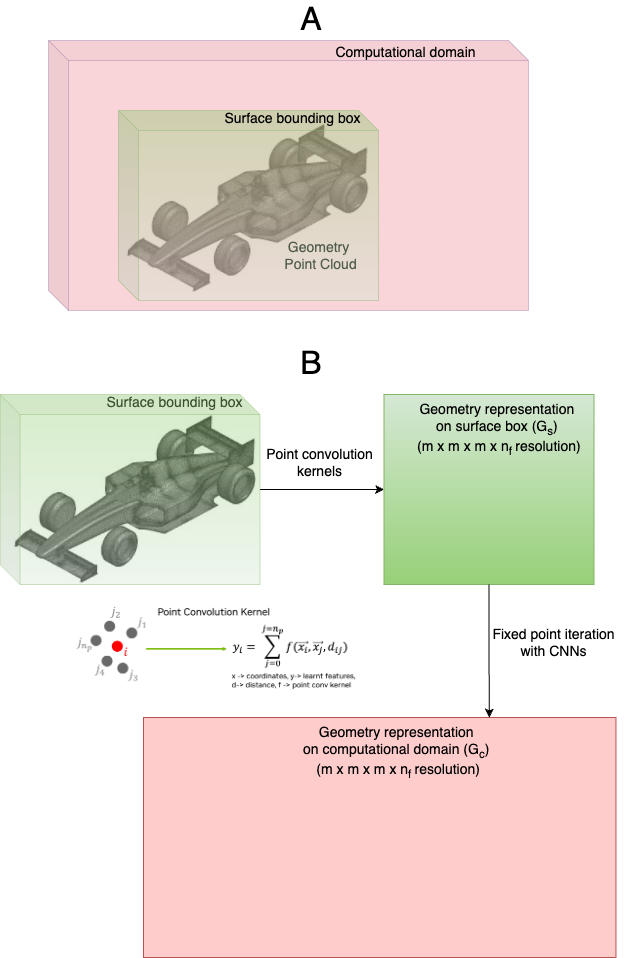}}
\caption{A: Bounding box representation and B: Geometry NN architecture}
\label{fig2}
\end{center}
\end{figure}

In this section, we describe the details of the geometry representation network used to calculate the global geometry representation from a point cloud. Fig. \ref{fig2} describes the various components of the geometry representation neural network. First, a tight-fitted bounding box is constructed around the surface in addition to a bounding box representing the computational domain as shown in Fig. \ref{fig2}A. The features of the geometry point cloud, such as spatial coordinates, are projected on to a N-D structured grid of resolution $m \times m \times m \times f$ overlaid on the surface bounding box using learnable point convolution kernels. The point convolution kernel used for projecting point clouds onto N-D structured grids is described in Eq. \ref{eq:pc}, where $\vec{x_i}$, $\vec{x_j}$ represent different sets of point clouds, $d_{ij}$ the distance metric between them and $f$ represents a fully connected deep neural network. In this work, the point convolution kernel is implemented using custom, GPU accelerated ball query layers using NVIDIA Warp. 
\begin{equation} 
    y_i = \sum_{j=0}^{j=n_y} f(\vec{x_i}, \vec{x_j}, d_{ij})
    \label{eq:pc} 
\end{equation}
The point convolution kernel depends on 2 additional factors, a radius of influence and number of points specified in the kernel ($n_y$). The radius of influence determines the physical size of the kernel and the extent of information learned from the neighbors. A larger radius signifies a bigger kernel and means that the geometry information propagates further into the domain. A smaller radius learns smaller kernels and captures finer geoemtry details. In the DoMINO architecture, we adopt a multi-scale approach by using a range of kernel sizes controlled by the radius parameter to capture both finer geometry features and long-range interactions of geometry. 

A structured grid of the same resolution is also constructed in the computational domain bounding box. Geometry features are propagated into the computational domain bounding box using 2 methods, 1) a separate set of multi-scale point convolution kernels are learned to project geometry information on to the computational domain grid and 2) the features on the surface bounding box grid ($G_s$) are propagated to the computational domain bounding box grid (($G_c$) using CNN blocks containing convolution, pooling, and unpooling layers. The CNN blocks are evaluated iteratively for a specified number of iterations. Although not explored in this work, a stopping criterion (such as $L_2 (G_s, G_c) < 1e^{-3}$) can be specified to break the iterations using a fixed-point iteration strategy. The $m \times m \times m \times f$ features calculated on the grid in the computational domain represent the geometry point cloud. In addition this, a SDF and its gradient components are also calculated on the computational domain grid and appended to the learned features to provide additional information about the topology of the geometry.

\subsection{Local geometry representation}
\label{local_encoding}

The local geometry representation depends on the physical location within the computational domain where the solutions fields are evaluated. Hence, before computing the local geometry representation, a batch of points is sampled in the computational domain. For each sampled point in the batch, $p$ neighboring points are sampled randomly around them to form a computational stencil of points similar to finite volume and element methods. Fig. \ref{fig1} represents a single sampled point and $p$ neighboring points randomly sampled around it. The local geometry representation is learned by drawing a subregion around the computational stencil of $p+1$ points. A point convolution kernel, as described in Eq. \ref{eq:pc}, is used to extract the local features of resolution, $l \times l \times l \times n_f$, in this subregion from the global geometry representation on the computational domain. The extracted local features are further transformed using fully connected neural networks. This local geometry representation is used to evaluate the solution fields on the sampled points using the aggregation network described Section \ref{aggregation_net}.

\subsection{Aggregation network}
\label{aggregation_net}

The local geometry representation represents the learned features of the geometry and solution in the vicinity of the computational stencil of the sampled points and its neighbors. Each of the points in this computational stencil are represented by their physical coordinates in the computational domain, SDF at these coordinates, normal vector from the center of mass of the domain, and surface normals (only if the point is on the surface). These input features are passed through a fully connected neural network, known as the basis function neural network, where a latent vector is computed representing these features for each of the points in the computational stencil. Each latent vector is concatenated with the local geometry encoding and passed through another set of fully-connected layers to predict a solution vector on each of the points in the computational stencil. The solution vector is averaged using an inverse distance weighing scheme to predict the final solution vector at the sampled point. A separate instance of the aggregation network is used for each solution variable but the local geometry representation across these remains the same.
\begin{figure}[ht!]
\begin{center}
\centerline{\includegraphics[width=0.8\columnwidth]{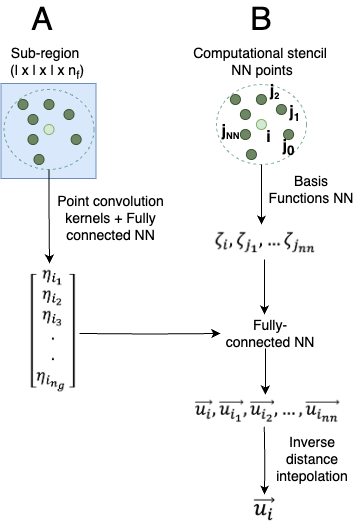}}
\caption{A: Local geometry encoding and B: Aggregation network}
\label{fig3}
\end{center}
\end{figure}

\section{Experiments details}
\label{experiments}

In this section, we demonstrate the application of DoMINO to predict key aerodynamic quantities on the surface of cars and in the volume around the cars. Most ML models are in literature exhibit a trade-off between scalability, accuracy and generalizability. ML models are typically trained on significantly down sampled meshes or point clouds. However, the proposed DoMINO model architecture provides a possible solution to simulation of large-scale aerodynamic simulations without sacrificing accuracy.

\subsection{Problem setup}
\label{problem}

In this study, the DoMINO model is trained to predict volume and surface fields in the computational domain. The volume fields considered are, velocity vector, $\vec{u}$, pressure, $p$ and turbulent viscosity, $\nu_t$ and the surface fields are pressure, $p$, and the wall-shear-stress vector, $\vec{\tau}$. The input to the model is an STL of the car geometry which is a triangulated representation of the surface. The coordinates of the nodes of the STL are used as input to learn the encoded geometry representation. In the aggregation network, the coordinates of the sampled points, SDF and the normal vector from the center of mass of the car geometry are considered as input features. When points are sampled on the surface of the car geometry, the surface normals at those points are also considered as part of the input features. Hidden representations learned from these input features, using neural networks, combined with the local encoded geometry representation are used to predict the respective fields on the sampled points. If the points are sampled on the surface, then the surface fields are predicted and compared with ground-truth values available on those sampled points to calculate loss. On the other hand, if the points are sampled in the volume around the car geometry then the volume fields are computed and compared with the corresponding ground truth values to calculate loss. The computational domain is trimmed for training the DoMINO model such that 2 car lengths in the downstream flow direction, 1 car length in the upstream direction and 1 car width and height in the other axial directions are considered. Finally, it is important to note that the simulation data is used as is to train the model and is not down-sampled or pre-processed in any manner. 

\subsection{Dataset}
\label{dataset}

\begin{figure}[ht!]
\begin{center}
\centerline{\includegraphics[width=1\columnwidth]{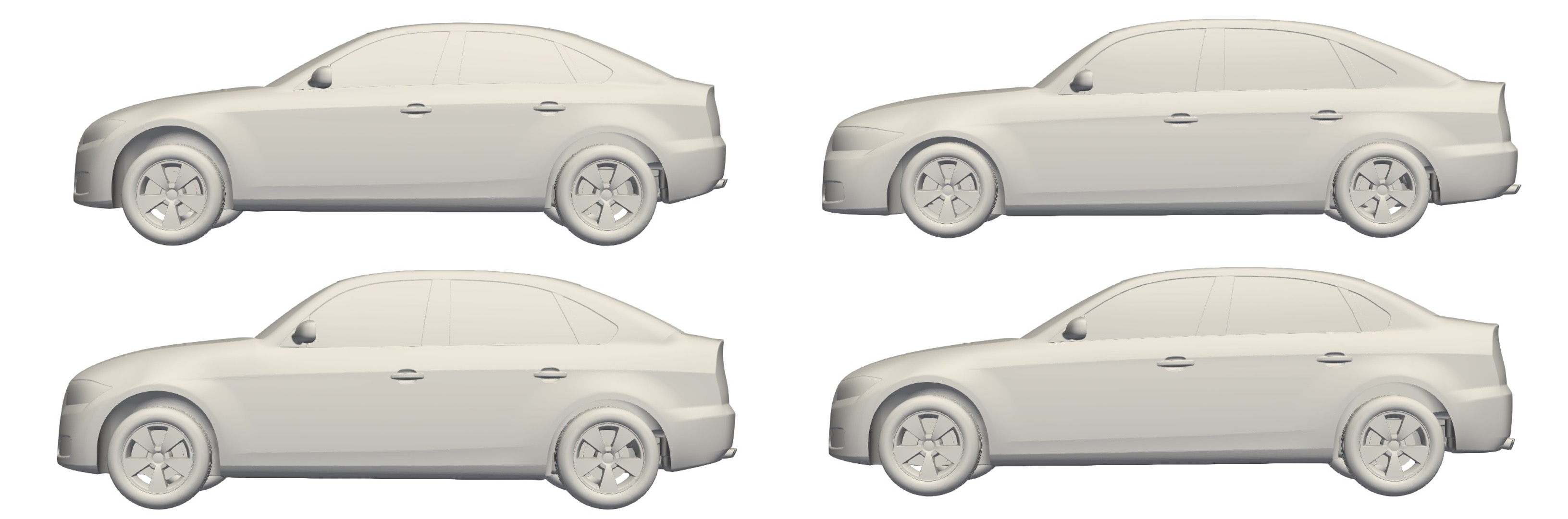}}
\caption{Examples of variation in geometry in simulation dataset}
\label{fig4}
\end{center}
\end{figure}

In this study, we use the DrivAerML dataset, a publicly available, high-fidelity dataset comprising of volume and surface data for 500 geometrically morphed variants of the DrivAer Notchback car \citep{ashton2024drivaerml}. Figure \ref{fig4} shows different examples of the car geometry variations performed in this dataset. The simulations in this dataset were carried out using the hybrid RANS/LES scale-resolving method. The time-averaged data is stored in the form of VTP and VTU files for each geometry variant. The VTP files contain the surface fields, namely pressure and wall-shear-stresses whereas the VTU files contain the volume fields, namely velocity, pressure and turbulent viscosity. The volume meshes across the 500 simulations cases are of the order of 150 million elements and the surface meshes are of the order 10 million elements. We split the simulations into a train and test set based on the drag force trends. 10\% of the samples are reserved in the test set with about 20\% of the test set consisting of out-of-distribution samples based on the ranges of the drag force. The out-of-distribution samples contain the cases with some of the lowest and the highest drag forces and are not seen by the model during training.  
\subsection{Model training}
\label{model training}
The DoMINO model is trained using the Adam optimizer. A reduce-on-plateau learning rate scheduler is used such that the initial learning rate, $1e^{-3}$, progressively drops down to a final learning rate, $1e^{-6}$, with the training epochs. The model is trained for a total of 500 epochs. Training is performed in float-16 precision format using Automatic Mixed Precision (AMP). The loss function used for training includes a mean-squared-error loss calculated on the sampled mesh points evaluated by the model on the surface of the car geometry and in the volume around it. The loss is calculated separately on each solution variable and then added together. In addition a mean-squared-loss is calculated for the area-weighted surface predictions for each surface variable and added to the overall loss formulation. During each epoch of training, the loss is calculated only on a subset of points randomly sampled from the simulation mesh and a new set is sampled each time. An area-weighted sampling algorithm is used for the sampling points on the surface, whereas a random uniform sampling algorithm is used for the volume points. Similarly, nodes of the STL are also sampled using a random uniform sampling algorithm for learning the geometry encoding. The number of points that can be sampled every epoch is dependent on the GPU memory availability. Additional details related to the model hyperparameters and training routine can be found in the Modulus Github repo \cite{contributors2023nvidia}.

\section{Results and analysis} 
\label{results}

\begin{figure}[ht!]
\begin{center}
\centerline{\includegraphics[width=0.6\columnwidth]{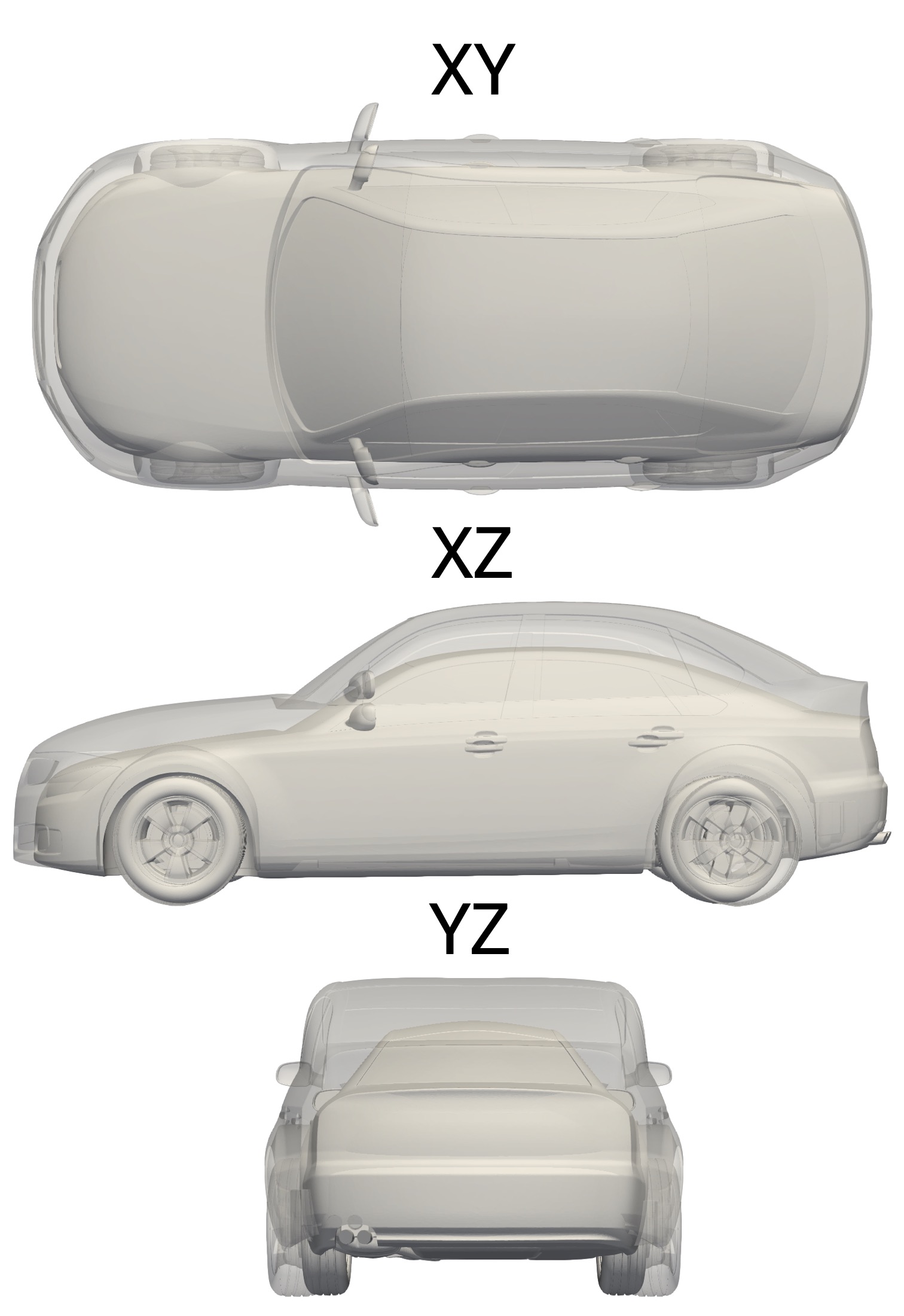}}
\caption{Out-of-distribution test samples}
\label{fig4:1}
\end{center}
\end{figure}

The trained DoMINO model is validated on a test set across several metrics for both the surface and volume quantities. The test set consists of both in-distribution and out-of-distribution samples, and the categorization is carried out based on the minimum and maximum value of the drag force in the training set. For example, if a test sample has a drag force that falls within the min-max range of the training set, then it is considered an in-distribution sample and if outside, then it is an out-of-distribution sample. In addition to the average error metrics and design trends, in this section, we will also present the contour and line plots for both surface and volume quantities for 2 out-of-distribution test samples, ids $419$ and $439$ from the dataset. These samples correspond to designs resulting the smallest and largest drag force in the complete DrivAerML dataset and are not a part of the training set. It may be observed from Fig. \ref{fig4:1} that there is substantial geometric variation between these samples resulting in around 300 N difference in the calculated drag force.

\subsection{Surface quantities}
\label{surface_results}

\begin{figure*}[ht!]
    \centering
    \begin{subfigure}[b]{0.49\linewidth}
      \centering
      \includegraphics[width=\linewidth]{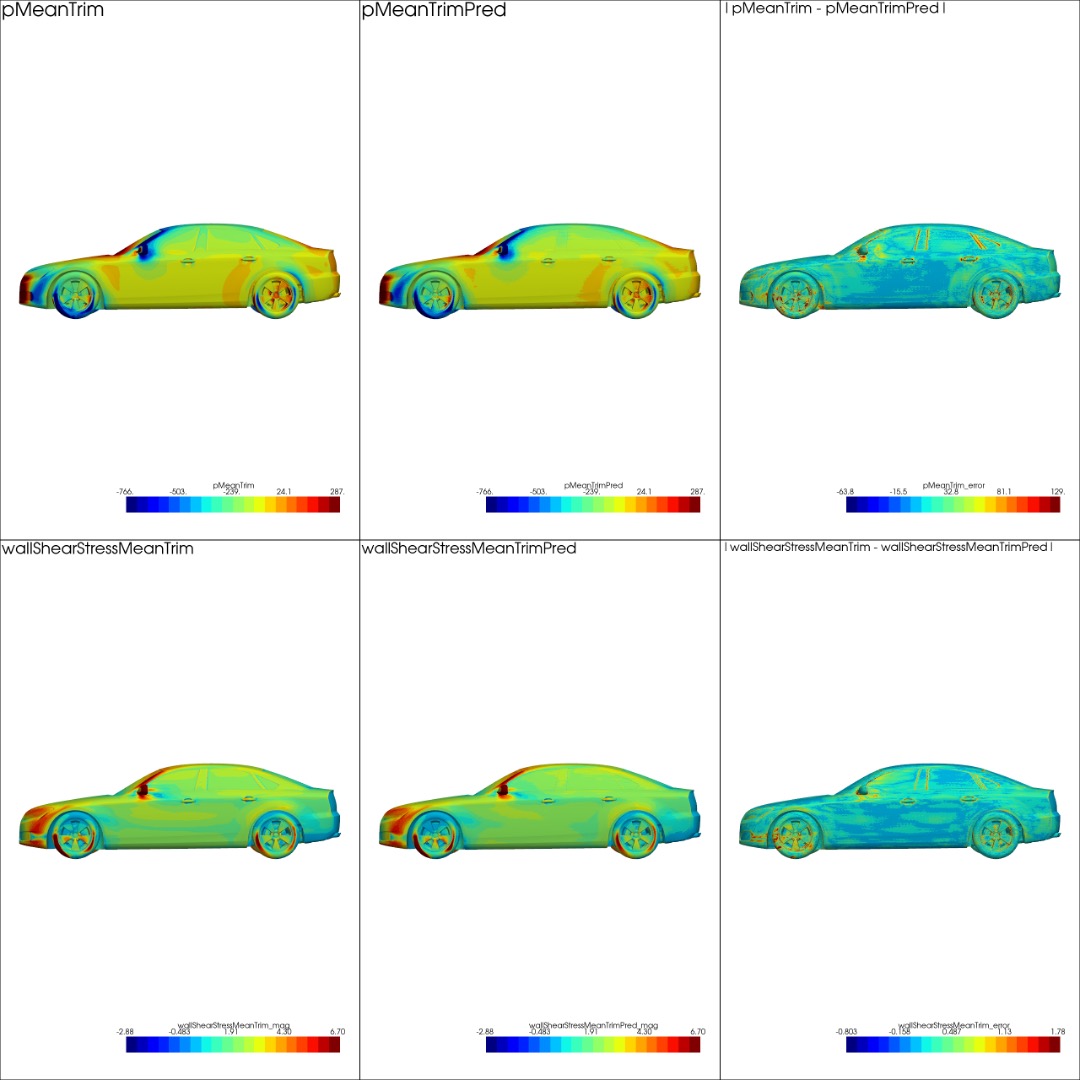}
      \caption{Surface contours for test sample id 419}
      \label{fig:sub1}
    \end{subfigure}
    \hfill
    \begin{subfigure}[b]{0.49\linewidth}
      \centering
      \includegraphics[width=\linewidth]{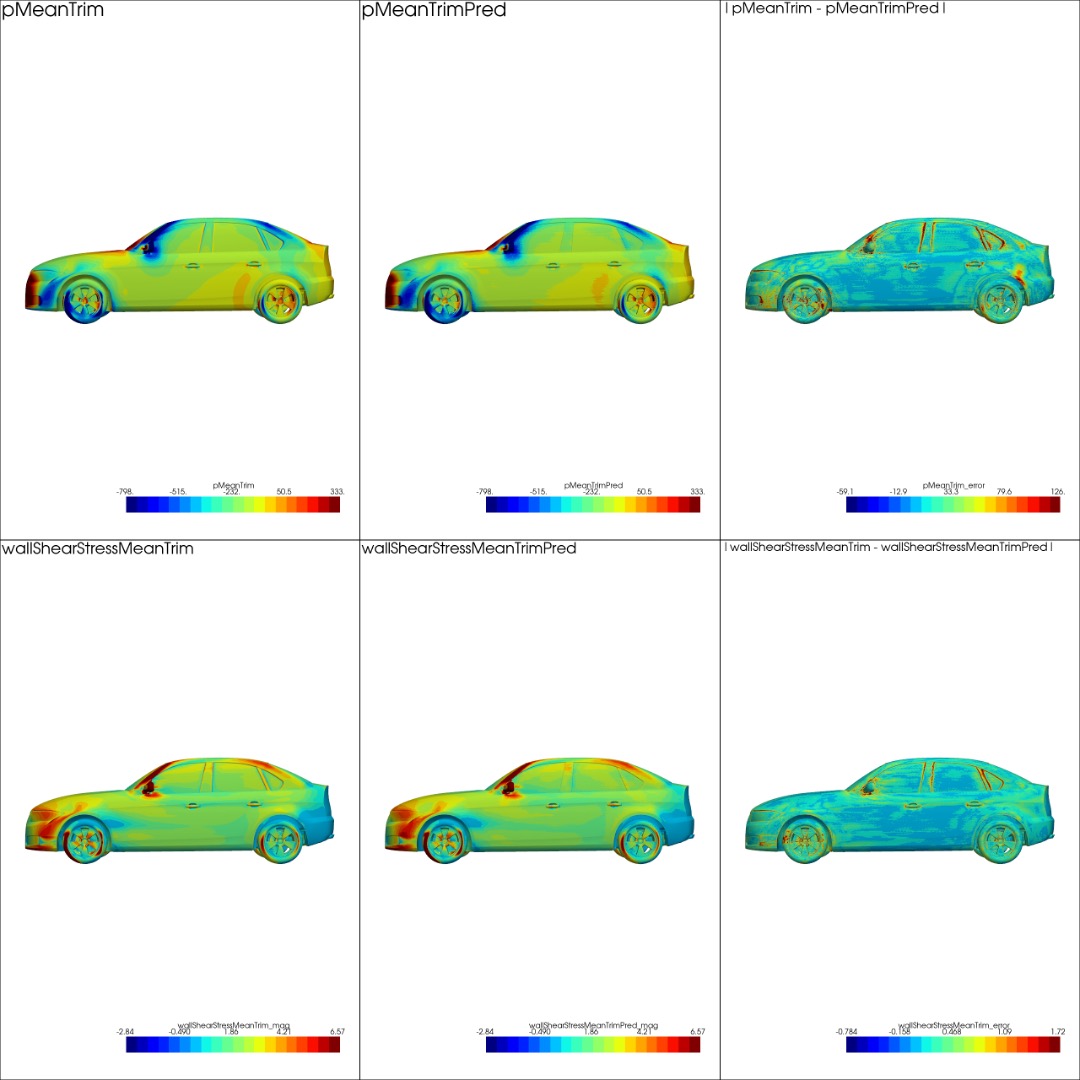}
      \caption{Surface contours for test sample id 439}
      \label{fig:sub2}
    \end{subfigure}
    \caption{Surface contour comparisons}
    \label{fig5}
\end{figure*}
First we present the results on the prediction of surface quantities. In Table \ref{surface-table} we show the relative $l_2$ and area weighted relative $l_2$ errors for pressure and wall-shear stress fields calculated on the surface of the car and averaged over all the test cases. The relative $l_2$ errors are calculated using Eq. \ref{eq:l2}.
\begin{equation} 
    \epsilon = \frac{\sqrt{\sum y_T^2 - y_P^2}}{\sqrt{\sum y_T^2}}
    \label{eq:l2} 
\end{equation}
where, $y_T$ and $y_P$ correspond to the true and predicted fields on cell centers of surface elements. For area weighted errors the field values are multiplied by the corresponding facet areas of the elements. The area-weighted relative errors are lower for all solution variables indicating that most of the errors occur on surface facets with small areas. The relative errors for Y and Z wall-shear-stresses are comparatively worse than pressure and X wall-shear-stress because of their small magnitudes. 
\begin{table}[ht!]
\caption{Average surface error metrics on test set}
\label{surface-table}
\vskip 0.15in
\begin{center}
\begin{small}
\begin{sc}
\begin{tabular}{lccr}
\toprule
Field & Rel. L-2 & Area-weighted rel. L-2 \\
\midrule
Pressure & 0.1505& 0.1181\\
X-Wall-Shear & 0.2124& 0.1279\\
Y-Wall-Shear & 0.302& 0.2769\\
Z-Wall-Shear & 0.3359& 0.229\\
\bottomrule
\end{tabular}
\end{sc}
\end{small}
\end{center}
\vskip -0.1in
\end{table}
In Fig. \ref{fig5} we present the surface contour comparisons between simulated results and DoMINO predictions and the error between them. It maybe observed that both pressure and wall-shear-stress magnitude are captured reasonably well on different surfaces of both the test designs including the windshield, side mirrors, underbody etc.
\begin{figure*}[ht!]
\begin{center}
\centerline{\includegraphics[width=0.6\linewidth]{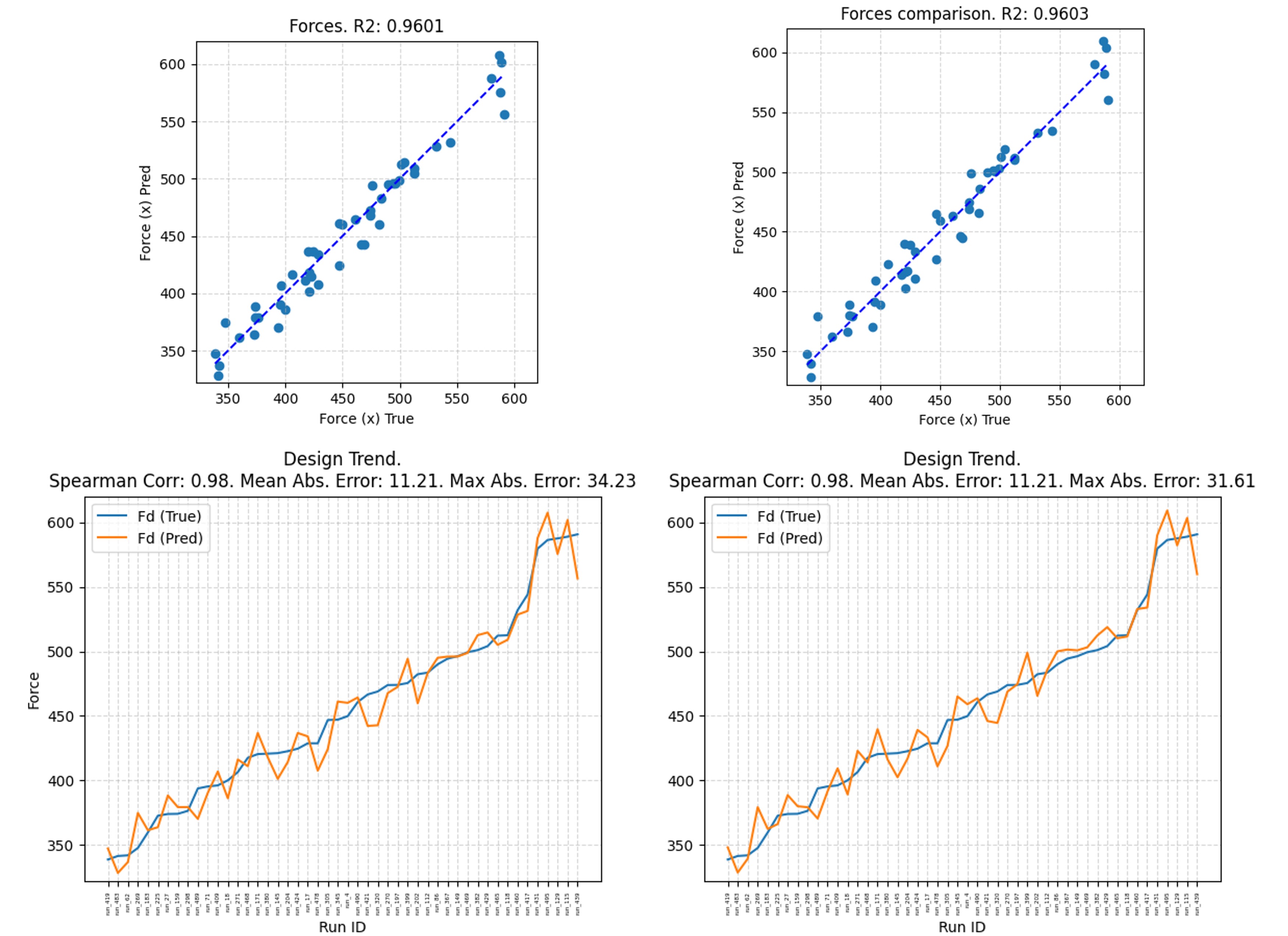}}
\caption{Drag force design trends and regression, Left: predicted using a simulation surface mesh and Right: predicted using a uniform surface point cloud}
\label{fig6}
\end{center}
\end{figure*}
Fig. \ref{fig6} shows the drag force design trends and the regression plot between the simulated and DoMINO predicted drag force for the various test designs. The dashed blue line in the regression plot represents the ideal case where, predictions match the true values perfectly. We conduct this experiment with 2 scenarios, 1) solution fields, pressure and x-wall-shear-stress are evaluated on the surface mesh and integrated to calculate the drag force and 2) solution fields, pressure and x-wall-shear-stress are evaluated on a uniform point cloud with 10 million points generated on the surface from the STL representation and integrated to calculate the drag fore. Each point in the uniform point cloud has the same surface area and the normals are derived from the STL facets. In both the scenarios, it may be observed that the coefficient of determination $R_2$ is 0.96 indicating a reasonable match between simulated drag force and DoMINO predictions. The design trends are calculated by arranging the drag forces of the test designs in an ascending order and comparing them against the drag force predictions for the same designs. The design trends are captured well, however smaller directional changes between successive designs shows some oscillatory behavior in the DoMINO predictions and resolving this will be a focus of future efforts. Finally, this experiment shows the invariance of the DoMINO model architecture to the spatial distribution of points. In both scenarios, the predicted regression coefficient and the drag force trends match very closely. The independence and insensitivity between the model and the spatial distribution of evaluation points is a an extremely important feature of this architecture especially for modeling engineering simulations where meshing can be expensive. 

\subsection{Volume quantities}
\label{volume_results}

\begin{figure*}[ht!]
    \centering
    \begin{subfigure}[b]{0.7\linewidth}
      \centering
      \includegraphics[width=\linewidth]{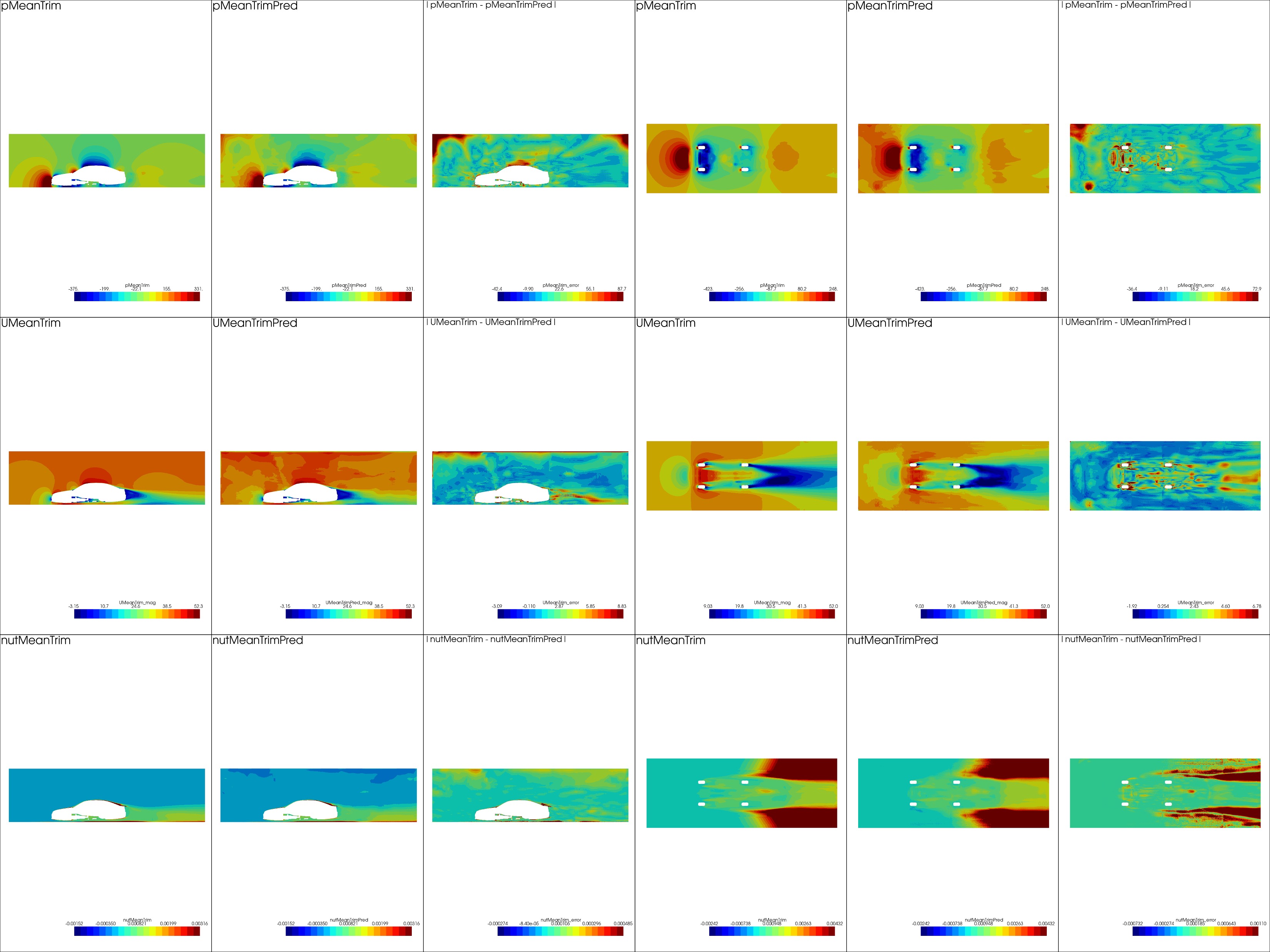}
      \caption{Volume contours for test sample id 419}
      \label{fig:vsub1}
    \end{subfigure}
    \hfill
    \begin{subfigure}[b]{0.7\linewidth}
      \centering
      \includegraphics[width=\linewidth]{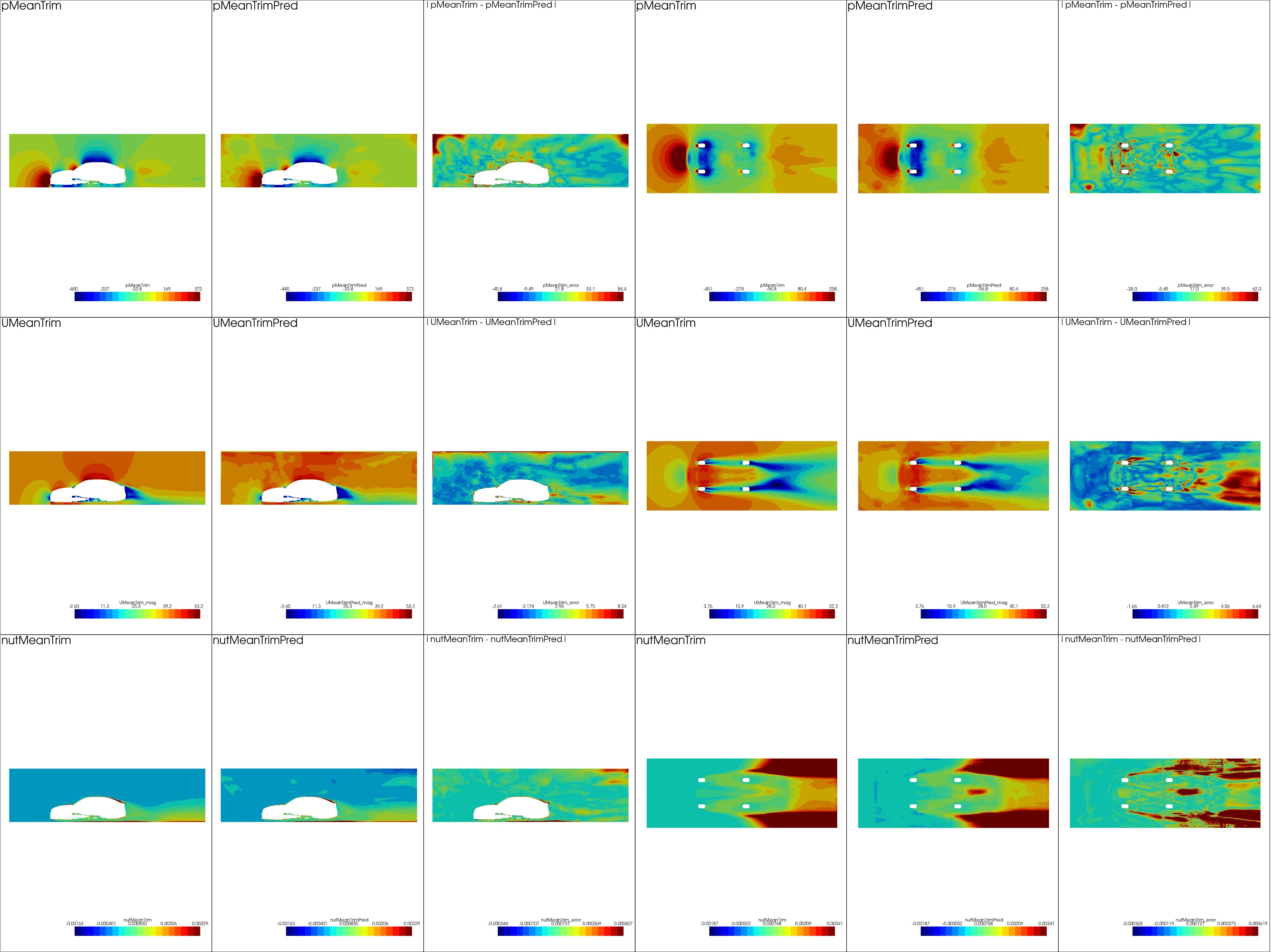}
      \caption{Volume contours for test sample id 439}
      \label{fig:vsub2}
    \end{subfigure}
    \caption{Volume contour comparisons on X-Y and X-Z planes}
    \label{fig7}
\end{figure*}
\begin{figure*}[ht!]
\begin{center}
\centerline{\includegraphics[width=1\linewidth]{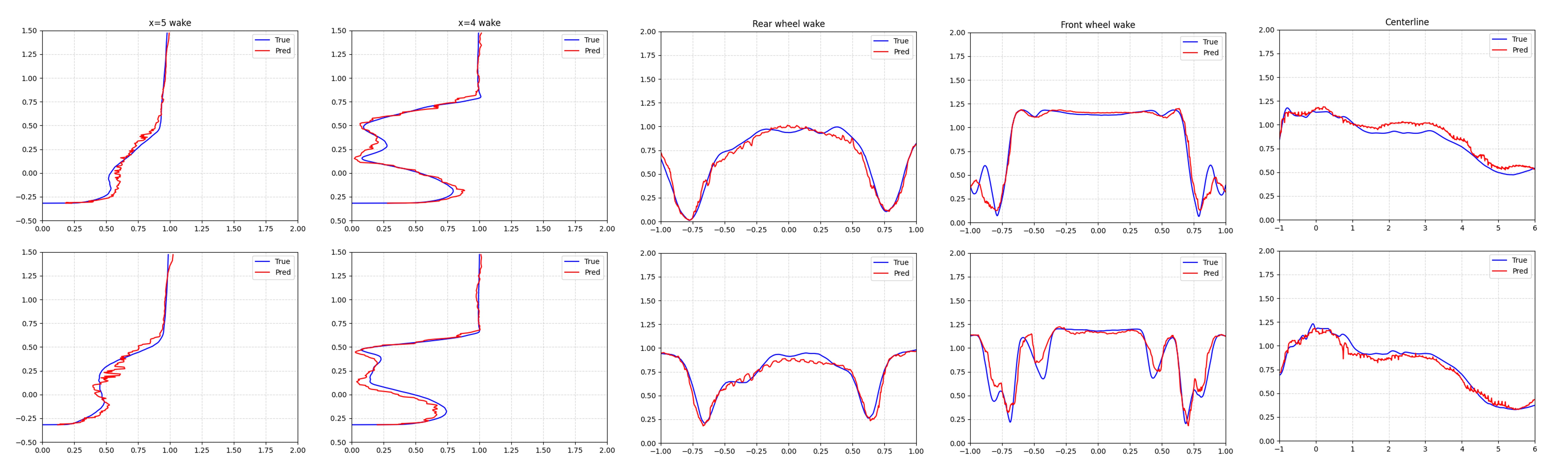}}
\caption{Velocity line plots at different locations (Sample 419 (top) vs Sample 439 (bottom)}
\label{fig8}
\end{center}
\end{figure*}
\begin{table}[ht!]
\caption{Average volume error metrics on test set}
\label{volume-table}
\vskip 0.15in
\begin{center}
\begin{small}
\begin{sc}
\begin{tabular}{lcr}
\toprule
Field & Rel. L-2 \\
\midrule
Pressure  & 0.2193\\
X-velocity & 0.2397\\
Y-velocity & 0.5025\\
Z-velocity & 0.4567\\
Turb-viscosity & 0.2175\\
\bottomrule
\end{tabular}
\end{sc}
\end{small}
\end{center}
\vskip -0.1in
\end{table}
Next, we present the volumetric results obtained from the DoMINO model evaluated in a bounding box around the car. Table \ref{volume-table} shows the relative $l_2$ errors, as described in \ref{eq:l2}, observed between DoMINO predictions and the simulated results averaged over the test set for different volume variables including, velocity, pressure and turbulent viscosity. A higher relative error is observed for Y and Z velocity components due to their sparse distribution and lower magnitudes in large parts of the computational domain.
Fig \ref{fig7} shows the comparisons between simulated results and DoMINO predictions for the pressure, velocity magnitude and turbulent-viscosity fields plotted on the X-Z and X-Y planes. The X-Y plane is cut through the center of the car (z = 0) while the X-Z plane represents the flow dynamics in the region between the floor and the car underbody. 
Additionally in Fig. \ref{fig8}, we plot the predicted and simulated velocities at 5 locations, in the wake at downstream distances of x=4 and x=5, on the rear and front wheel and along the centerline through the car. It can be observed from the contours as well as the line plots that the predicted fields capture the flow dynamics reasonably accurately, especially in regions of the wake, underbody, and near the hood of the car. Moreover, sparse fields such as turbulent viscosity are also modeled accurately in the wake downstream of the car. The contours for all fields show an odd behavior near the edges of the bounding box. Although this behavior does not affect the important regions near the car where the flow field predictions are reasonably accurate, the reason for it is not clearly understood and will be investigated further.  

\section{Conclusion}

In this paper, we introduced a new ML-architecture DoMINO, designed to address key challenges in the use of ML for large-scale numerical simulations. The model has several unique features that enable accurate modeling of large-scale numerical simulations.

\begin{enumerate}
    \item The model operates directly on point cloud representations of geometries. It uses dynamic ball query kernels implemented using NVIDIA Warp to represent geometry point clouds into global representations of geometry. NVIDIA Warp provides significant GPU acceleration as compared to using other PyTorch based alternatives. Moreover, this makes the model very flexible in utility as it does not require generation of a mesh during inference as often times mesh generation is an expensive process for large-scale industrial simulations.
    \item A range of kernel sizes are used to propagate the geometry information into the computational domain using an iterative process thereby enabling efficient modeling of both short- and long-range interactions.
    \item The global geometry representations are dense and high-dimensional and cannot be used as is to accurately predict solutions in all regions of the computational domain. As a result, local geometry encodings are extracted using dynamic ball query points around the points where the solution is calculated. Additionally as these local geometry representations are learned during training they extract the essential information from the geometry required to accurately predict solution fields in different regions of the computational domain. For example, in predicting the flow fields in the wake, only the local geometry features required for learning these are extracted from the global geometry encoding during training. The local geometry encoding also improves the generality of the model to different geometries.
    \item The solution at any point in the computational domain is heavily influenced by the local stencil of points around it. A local stencil of points, similar to a finite volume or element method, is constructed around each sampled point, processed using a basis function neural network and aggregated to calculate the solutions on the sampled points. This enables capturing local information important for accurate learning of the solution fields.
    \item Finally, since the model can predict solutions on arbitrarily sampled points, it is not dependent on how these points are sampled. For example, the sampled points may be the nodes of a simulation mesh or randomly sampled in the computational domain. Additionally, the model evaluation is not limited in memory or compute by the number of sampled points as the evaluation can happen in batches. As a result, the DoMINO model can easily scale to large computational domains and meshes.
\end{enumerate}

The DoMINO model is demonstrated on the external automotive aerodynamics use case, which is challenging because of the large-scale nature of the simulations (meshes of the order of hundreds of millions to billions), difficulty in representing finer features of car geometries and accurately modeling both surface and volume field quantities. The experiments show that the model can accurately capture both the flow fields on the volume and surface, as well as other engineering metrics such as drag force regression and design trends. We also show the generalization of the model to prediction on out-of-distribution test samples and different mesh or point-cloud configurations. Finally, the experiments also showcase the model's ability to scale to large meshes and point clouds with real time inference making it suitable for other engineering applications.

Future work will focus on extending the model architecture to improving accuracy and performance, especially in resolving the oscillatory predictions using smoothing constraints in the design trends and line comparisons. We will also explore the impact of adding fixed point iterations in our iterative learning strategy. The convergence of the fixed point iteration can depend on the ratio of the physical size of the surface bounding box and computational domain bounding box. If this ratio is too small then a multi-layered approach can be used by adding bounding boxes of different sizes between the surface and computational domain and solving fixed point iterations between each transformation similar to a multi-grid approach. Currently, the model relies on fixed resolution latent spaces but in the future we would like to add multi-resolution latent spaces, in an attempt to better capture the finer features in the geometry as well as further improve the long-range interactions. Additionally, we will explore the effect of hybrid training pipelines involving constraining the model training with PDE based losses computed using automatic differentiation. Finally, we would like to extend the model to transient problems and other large-scale applications in engineering simulation.


\bibliography{domino_paper}
\bibliographystyle{icml2025}

\end{document}